\documentclass[conference, a4, 11pt, twoside, onecolumn]{IEEEtran}

\usepackage{newtxmath}
\usepackage{amsmath}
\usepackage{amsfonts}
\usepackage{mathtools, nccmath}

\usepackage[superscript, noadjust]{cite}

\usepackage{graphicx}
\graphicspath{{figures/}}
\DeclareGraphicsExtensions{.pdf,.jpeg,.png,.eps}

\usepackage{booktabs}
\usepackage{array}
\usepackage{footnote}
\usepackage[acronym]{glossaries}
\usepackage{multicol}
\usepackage{color}
\usepackage{enumitem}
\usepackage{stfloats}
\usepackage[hidelinks]{hyperref}
\usepackage[protrusion, expansion, kerning, spacing]{microtype}

\usepackage{siunitx}
\newcommand{\beginsupplement}{%
        \setcounter{table}{0}
        \renewcommand{\thetable}{S\arabic{table}}%
        \setcounter{figure}{0}
        \renewcommand{\thefigure}{S\arabic{figure}}%
     }

\DeclarePairedDelimiter{\nint}\lfloor\rceil

\usepackage{geometry}
\newacronym{ebn}{EBN}{Efficient Balanced Network}
\newacronym{rnn}{RNN}{Recurrent Neural Network}
\newacronym{lif}{LIF}{Leaky Integrate and Fire}
\newacronym{snn}{SNN}{Spiking Neural Network}
\newacronym{dnn}{DNN}{Deep Neural Network}
\newacronym{stdp}{STDP}{Spike-Timing Dependent Plasticity}
\newacronym{bptt}{BPTT}{Back-Propagation Through Time}
\newacronym{mse}{MSE}{Mean-Squared Error}
\newacronym{iqr}{IQR}{Interquartile range}
\newacronym{lsm}{LSM}{Liquid State Machine}
\newacronym{ads}{ADS}{Arbitrary Dynamical System}
\newacronym{mac}{MAC}{Multiply-Accumulate}

\begin{document}

% -- Title
% Target: 20 words

% \title{Robust Classification on mixed-signal neuromorphic processors using Efficient Balanced Networks} %86 chars

%\title{Balanced Spiking Networks provide robustness to mismatch on mixed-signal neuromorphic devices} %83 chars

% \title{Arbitrary temporal signal classification in robust spiking neural networks} %66 chars

% \title{Supervised temporal signal classification in spiking neural networks that are robust against mismatch on neuromorphic processors} %128 chars

\title{Supervised training of spiking neural networks for robust deployment on mixed-signal neuromorphic processors} %108 chars, 14 words

% Author names and affiliations

% \and
% \IEEEauthorblockN{Dmitrii~Zendrikov}\\
% \IEEEauthorblockA{Institute of Neuroinformatics\\
% University of Zurich and ETH Zurich\\
% Winterthurerstrasse 190\\
% 8057 Zurich, Switzerland}
% \and
% \IEEEauthorblockN{Sergio~Solinas}\\
% \IEEEauthorblockA{Department of Biomedical Science\\
% University of Sassari\\
% Piazza Università, 21\\
% 07100 Sassari, Sardegna, Italy}
% \and
% \IEEEauthorblockN{Giacomo~Indiveri}\\
% \IEEEauthorblockA{Institute of Neuroinformatics\\
% University of Zurich and ETH Zurich\\
% Winterthurerstrasse 190\\
% 8057 Zurich, Switzerland}
% \and
% \IEEEauthorblockN{Dylan~R.~Muir}
% \IEEEauthorblockA{SynSense, Thurgauerstrasse 40\\
% 8050 Zurich, Switzerland}
% }

% Julian~B\"uchel$^{1,2}$,
%         Dmitrii~Zendrikov$^{2}$,
%         Sergio~Solinas$^{3,2}$,
%         Giacomo~Indiveri$^{1,2}$,
%         Dylan~R.~Muir$^{1,*}$
%         \thanks{$^1$SynSense, Thurgauerstrasse 40, 8050 Zurich, Switzerland; $^2$Institute of Neuroinformatics, University of Zurich and ETH Zurich, Winterthurerstrasse 190, 8057 Zurich, Switzerland; $^3$Department of Biomedical Science, University of Sassari, Piazza Università, 21, 07100 Sassari, Sardegna, Italy; *dylan.muir@synsense.ai}
% }
 
% \markboth{Journal of X,~Vol.~14, No.~8, August~2015}%
% {B\"uchel \MakeLowercase{\textit{et al.}}: Robust Classification using Efficient Balanced Networks}

\maketitle

% Authors (above the Abstract for Nature Sci Rep)
% \subsection*{Authors}
\vspace*{-6em}
\textbf{Julian~B\"uchel}; SynSense, Thurgauerstrasse 40, 8050 Zurich, Switzerland\\
\textbf{Dmitrii~Zendrikov}; Institute of Neuroinformatics,
University of Zurich and ETH Zurich, Winterthurerstrasse 190, 8057 Zurich, Switzerland\\
\textbf{Sergio~Solinas}; Department of Biomedical Science,  University of Sassari, Piazza Università, 21, 07100 Sassari, Sardegna, Italy\\
\textbf{Giacomo~Indiveri}; Institute of Neuroinformatics,
University of Zurich and ETH Zurich, Winterthurerstrasse 190, 8057 Zurich, Switzerland\\
\textbf{Dylan~R.~Muir}; SynSense, Thurgauerstrasse 40, 8050 Zurich, Switzerland\\
Corresponding author: Dylan R. Muir
\vspace{1em}

% Abstract
\begin{abstract}
\newpage
% Target: 150 words
% Current count: 206

% One or two sentences providing a basic introduction to the field, comprehensible to a scientist in any discipline
Mixed-signal analog/digital circuits emulate spiking neurons and synapses with extremely high energy efficiency, an approach known as ``neuromorphic engineering''.

% Two to three sentences of more detailed background, comprehensible to scientists in related disciples
However, analog circuits are sensitive to process-induced variation among transistors in a chip (``device mismatch'').
For neuromorphic implementation of \glspl{snn}, mismatch causes parameter variation between identically-configured neurons and synapses.
Each chip exhibits a different distribution of neural parameters, causing deployed networks to respond differently between chips.

% One sentence clearly stating the general problem being addressed by this particular study
Current solutions to mitigate mismatch based on per-chip calibration or on-chip learning entail increased design complexity, area and cost, making deployment of neuromorphic devices expensive and difficult.

% One sentence summarising the main result
Here we present a supervised learning approach that produces \glspl{snn} with high robustness to mismatch and other common sources of noise.

% Two or three sentences explaining what the main result reveals in direct comparison to what was thought to be the case previously, or how the main result adds to previous knowledge
Our method trains \glspl{snn} to perform temporal classification tasks by mimicking a pre-trained dynamical system, using a local learning rule from non-linear control theory.
We demonstrate our method on two tasks requiring memory, and measure the robustness of our approach to several forms of noise and mismatch.
We show that our approach is more robust than common alternatives for training \glspl{snn}.

% One or two sentences to put the results into a more general context
Our method provides robust deployment of pre-trained networks on mixed-signal neuromorphic hardware, without requiring per-device training or calibration.

\end{abstract}

% Note that keywords are not normally used for peer review papers.
% \begin{IEEEkeywords}
% Efficient Balanced Networks, Neural Computation, Neuromorphic Computing, Robust Classification
% \end{IEEEkeywords}

\IEEEpeerreviewmaketitle

% --- Main text
% Target: 5000 words
% Current count: 4420 words

\section*{Introduction}

% Motivation - mixed-signal HW is great for SNNs
Dedicated hardware implementations of \acrlong{snn}s (\glspl{snn}) are an extremely energy-efficient computational substrate on which to perform signal processing and machine learning inference tasks \cite{corradi, dynapse, Cassidy2016TrueNorthAH, neur_silicon_circ, class_using_dbns, schemmel_brainscales_hicann, loihi, spinnaker}. Optimal energy efficiency is achieved when using mixed-signal analog/digital neuron and synapse circuits following an approach known as ``neuromorphic engineering''\cite{Mead90neuromorphicelectronic}.
In these hardware devices, large arrays of neurons and synapses are physically instantiated in silicon, and coupled with flexible digital routing and interfacing logic in ``mixed-signal'' designs \cite{dynapse, schemmel_brainscales_hicann}.

% Mismatch exists, and creates parameter noise (other noise also exists)
However, all analog silicon circuits suffer from process variation across the surface of a chip, changing the operating characteristics of otherwise identical transistors — known as ``device mismatch'' \cite{Pelgrom_etal89,Tuinhout2009}.
In the case of spiking neurons implemented using analog or mixed-signal circuits, mismatch is expressed as parameter variation between neurons and synapses that are otherwise configured identically \cite{Qiao2016, mismatch_indi_sandam, neftci_indiveri, Aamir2018}.
The parameter mismatch on each device appears as frozen parameter noise, introducing variance between neurons and synapses in time constants, thresholds, and weight strength.

% Mismatch can be good, but raises problems for deployment
Parameter noise in mixed-signal neuromorphic devices can be exploited as a symmetry-breaking mechanism, especially for neural network architectures that rely on randomness and stochasticity as a computational mechanism \cite{sheik_mismatch, neur_extr_le_machines, Maass_etal_2002_lsm, eliasmith_2005, neckar}, or can be exploited to improve in-situ training of Bayesian networks via MCMC sampling\cite{dalgaty}.
However, random architectures can raise problems for commercial deployment of applications on  mixed-signal devices: the parameter noise would affect neuronal response dynamics, and these device to device variations could affect and degrade the system performance of individual chips. A possible solution would be to perform a post-production device calibration step or re-training, but this would raise deployment costs significantly and not scale well with large volumes.
In addition to device mismatch, mixed-signal neuromorphic systems also suffer from other sources of noise, such as thermal noise or quantisation noise introduced by restricting synaptic weights to a low bit-depth.

% Configuration problem for SNNs
In contrast to current mainstream Deep Neural Networks (\acrshort{dnn}s), spiking networks suffer from a severe configurability problem.
The backpropagation algorithm permits configuration of extremely deep NNs for arbitrary tasks \cite{deep_nets}, and is effective also for network models with temporal state \cite{BPTT}, but is difficult to apply to the discontinuous dynamics of \gls{snn}s \cite{delbruck_spikingBP, neftci_bp, Bellec2020}.
Methods to approximate the gradient calculations by using surrogate functions\cite{neftci_surrogate}, eligibility traces\cite{three_factor} or adjoint networks\cite{wunderlich2020eventprop} have provided a way to adapt backpropagation for spiking networks.
Non-local information is required for strict implementation of the backpropagation algorithm, but random feedback\cite{Lillicrap2016} and local losses\cite{decolle} have been employed with some success to train multi-layer spiking networks.
Alternative approaches using initial random dynamics coupled with error feedback and spike-based learning rules can permit recurrent \gls{snn}s to mimic a teacher dynamical system\cite{Nicola2017, Gilra_2017}.
Strictly-local spike-timing-based learning rules, inspired by results in experimental neuroscience\cite{Markram2012}, have been implemented in digital and mixed-signal neuromorphic devices, as they provide a better match to the distribution of information across neuromorphic chips\cite{fusi}.
Unfortunately, local spike-dependent rules such as \gls{stdp} are themselves not able to provide supervised training of arbitrary tasks, since they do not permit error feedback or error-based modification of parameters.
In both cases, implementing strictly local or backpropagation-based learning infrastructure on-chip adds considerable complexity, size and therefore cost to neuromorphic hardware designs.
This cost makes it impractical to use on-chip learning and adaptation to solve the mismatch problem on mixed-signal architectures.

% Noise can be combated through network architecture
Robustness to noise and variability can be approached from the architectural side.
For example, a network architecture search approach can identify networks that are essentially agnostic to precise weight values\cite{gaier2019weight}.
However, these networks rely on complex combinations of transfer functions which do not map to neuromorphic \gls{snn} designs.

Alternatively, a class of analytically-derived network architectures have been proposed for spiking networks, known as Efficient Balanced Networks\cite{bourdoukan_etal_2012, boerlin_etal_2013, Deneve2016, deneve_etal_2017, alemi_etal_2018_learning, pmid32176682, Calaim2020.06.15.148338}, relying on a balance between excitation and inhibition to provide robustness to sources of noise including spike-time stochasticity and neuron deletion.
These networks can be derived to mimic an arbitrary linear dynamical system through an auto-encoding architecture\cite{boerlin_etal_2013} or can learn to represent and mimic dynamical systems\cite{bourdoukan_etal_2012, alemi_etal_2018_learning, deneve_etal_2017, pmid32176682}.
We propose to adapt the learning machinery of this spiking architecture to produce deployable \gls{snn}-based solutions for arbitrary supervised tasks that are robust to noise and device mismatch.

% Paper overview
In this work we present a method for training robust networks of \gls{lif} spiking neurons that can solve supervised temporal signal regression and classification tasks.
We adopt a knowledge distillation approach, by first training a non-spiking \gls{rnn} to solve the desired supervised task using \gls{bptt}\cite{BPTT}.
By then interpreting the activations of the \gls{rnn} as a teacher dynamical system, we train an \gls{snn} using an adaptation of the learning rule from ref.~\citenum{alemi_etal_2018_learning} to mimic the \gls{rnn}.
We show that the resulting trained \gls{snn} is robust to multiple forms of noise, including simulated device mismatch, making our approach feasible for deployment on to mixed-signal devices without post-deployment calibration or learning.
We compare our method with several other standard approaches for configuring \gls{snn}s, and show that ours is more robust to device mismatch.

\section*{Results}
\begin{figure}
\centering
  \includegraphics[width=\columnwidth]{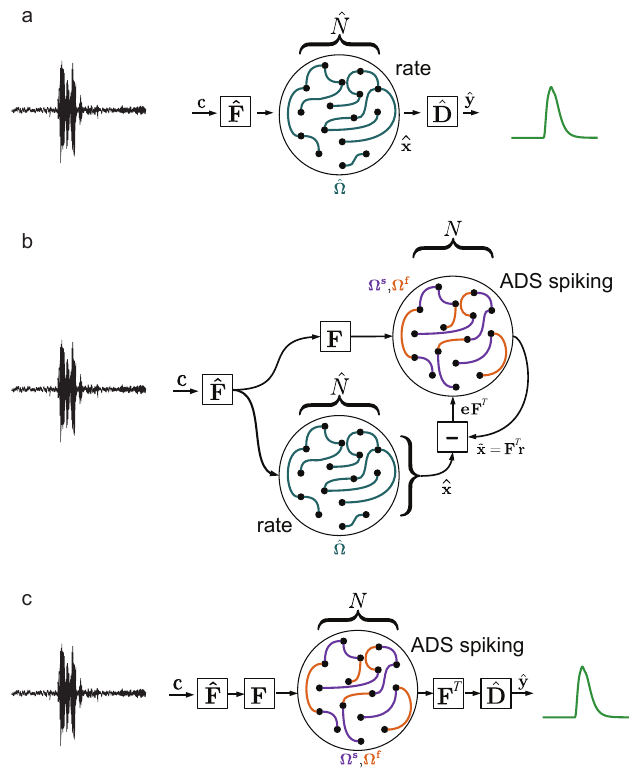}
  \caption{
    \textbf{Schematic overview of our supervised training approach.}
    \textbf{a} A recurrent non-spiking neural network with $\hat{N}$ neurons (``rate'') is trained using \gls{bptt} or a similar approach to implement the mapping $\textbf{c} \rightarrow \hat{\textbf{y}}$, via encoding and decoding weights $\hat{\textbf{F}}$ and $\hat{\textbf{D}}$, using the recurrent weights $\hat{\Omega}$ and resulting in the internal temporal representation of neural activity $\hat{\textbf{x}}$.
    \textbf{b} To train a robust spiking network for the task, a network with $N \neq \hat{N}$ LIF neurons (``ADS spiking'') is initialised with fast balanced feedback connections $\Omega^\textbf{f}$, analytically determined from a randomly chosen encoding matrix $\textbf{F}$. The ADS spiking network learns to represent the target signals $\hat{\textbf{x}}$ with reference to an error signal $\textbf{e} = \tilde{\textbf{x}} - \hat{\textbf{x}}$, by adapting slow feedback connections $\Omega^\textbf{s}$.
    \textbf{c} For inference, the ADS spiking network replaces the non-spiking rate network, and uses the encoding and decoding weights $\hat{\textbf{F}}$ and $\hat{\textbf{D}}$ to implement the trained task mapping $\textbf{c} \rightarrow \hat{\textbf{y}}$.
  }
  \label{fig:setup_complete}
\end{figure}

% General task overview
We assume a family of tasks defined by mappings $\textbf{c}(t) \rightarrow \hat{\textbf{y}}(t)$, where $\textbf{c}(t)\in \mathbb{R}^{d1}$ and $\hat{\textbf{y}}(t)\in \mathbb{R}^{d2}$ are temporal signals with arbitrary dimensionality (Figure~\ref{fig:setup_complete}a; see Methods).
For simplicity of notation we do not write the temporal dependency ``$(t)$'' for the remainder of the paper.
This definition encompasses any form of deterministic temporal signal processing or classification task without loss of generality. We refer to our network architecture as ADS (\textbf{A}rbitrary \textbf{D}ynamical \textbf{S}ystem) spiking networks.

% Method figure for training spiking ADS networks

% Brief method for training spiking ADS network
Our approach begins by training a non-spiking rate network to implement the arbitrary task mapping by learning the dynamical system
\begin{IEEEeqnarray*}{rCl}
    \tau\dot{\hat{\textbf{x}}}  & = & \hat{\Omega} f(\hat{\textbf{x}}) + \hat{\textbf{F}}\textbf{c} + b\\
    \hat{\textbf{y}}            & = & \hat{\textbf{D}}\hat{\textbf{x}}
\end{IEEEeqnarray*}
through modification of the recurrent weights $\hat{\Omega}\in\mathbb{R}^{\hat{N}\times\hat{N}}$; encoding and decoding weights $\hat{\textbf{F}}\in\mathbb{R}^{d1\times\hat{N}}$ and $\hat{\textbf{D}}\in\mathbb{R}^{\hat{N}\times d2}$; biases $b\in\mathbb{R}^{\hat{N}}$; time constants~$\tau\in\mathbb{R}^{\hat{N}}$; and non-linear transfer function $f(\cdot)=\tanh(\cdot)$.
\gls{bptt} or any other suitable approach can be used to obtain the trained rate network.

We subsequently train a network of spiking neurons to emulate $\hat{\textbf{x}}$, with leaky membrane dynamics defined by
$$\dot{V} = -\lambda V + \hat{\textbf{F}}\textbf{F}\textbf{c} - \Omega^{\textbf{f}}\textbf{o} + \Omega^{\textbf{s}}\textbf{o} + k\textbf{F}^T\textbf{e}$$
with spike trains $\textbf{o} = V>V_\textrm{thresh}$ produced when exceeding threshold voltages $V_\textrm{thresh}$; leak rate $\lambda$; and fast and slow recurrent weights $\Omega^{\textbf{f}}$ and $\Omega^{\textbf{s}}$ (Figure~\ref{fig:setup_complete}b; see Methods).
The decoded dynamics $\tilde{\textbf{x}} \approx \hat{\textbf{x}}$ are obtained from the filtered spiking activity $\textbf{r}$ with $\tilde{\textbf{x}} = \textbf{F}\textbf{r}$.
By feeding back an error signal $\textbf{e} = \tilde{\textbf{x}} - \hat{\textbf{x}}$ under the control of a decaying feedback rate $k$, the spiking network is forced to remain close to the desired target dynamics.
$\Omega^{\textbf{f}}$ is initialised to provide fast balanced feedback\cite{Deneve2016}, and $\Omega^{\textbf{s}}$ is learned using the rule
$$\dot\Omega^{\textbf{s}} = \eta \textbf{r}\textbf{F}\textbf{e}$$
under learning rate $\eta$ (see Methods and ref.~\citenum{alemi_etal_2018_learning}).
Note that we do not require complex multi-compartmental neurons or dendritic nonlinearities in our neuron model, but use a simple leaky integrate-and-fire neuron that is compatible with compact mixed-signal neuromorphic implementation\cite{dynapse}.
Once the spiking network has learned to represent $\tilde{\textbf{x}} \approx \hat{\textbf{x}}$ with high accuracy, we replace the rate network entirely with the spiking network (Figure~\ref{fig:setup_complete}c).

% Our method is able to learn a simple nonlinear temporal task
\subsection*{Temporal XOR task}
\begin{figure}
  \includegraphics[width=\columnwidth]{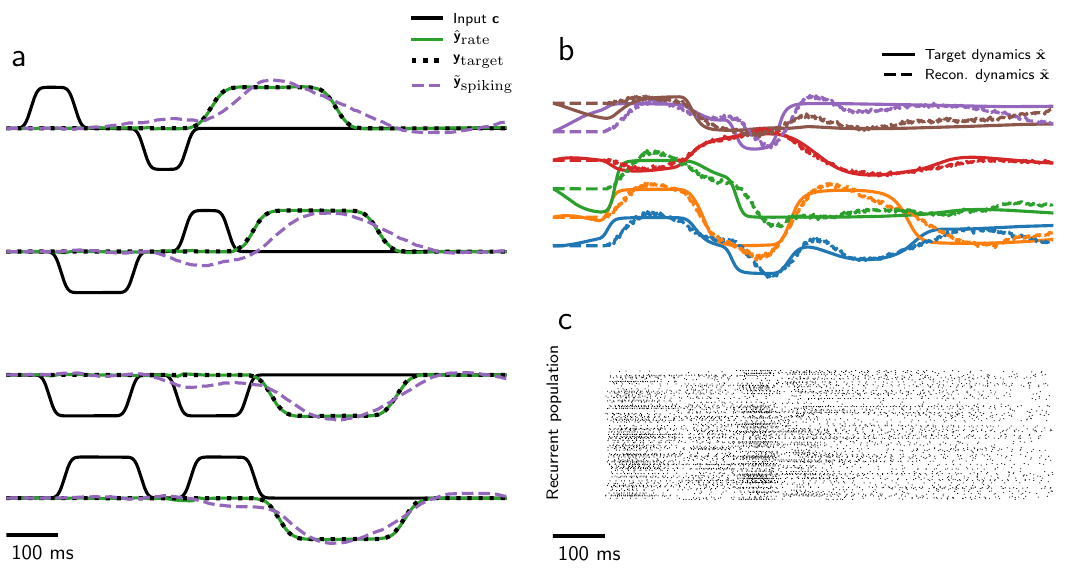}
  \centering
  \caption{
    \textbf{Our approach implements a supervised temporal non-linear sequence classification task to high accuracy.}
    \textbf{a} The non-spiking \gls{rnn} (green; $\hat{N} = 64$ neurons) is trained to perform a temporal XOR of the input (black), closely matching the target function (dotted). A spiking ADS network (dashed; $N = 320$ neurons) is trained to perform the same task.
    \textbf{b} The first six internal dynamical variables $\hat{\textbf{x}}$ of the \gls{rnn} are shown (solid), along with their reconstructed equivalents $\tilde{\textbf{x}}$ from the spiking ADS network (dashed).
    \textbf{c} The spiking activity of the ADS network.
    Panels \textbf{b} and \textbf{c} correspond to the first example in \textbf{a}.}
  \label{fig:temporal_xor}
\end{figure}

We begin by demonstrating our method using a nonlinear temporal XOR task (Figure~\ref{fig:temporal_xor}; see Methods).
This task requires memory of past inputs to produce a delayed output, as well as a nonlinear mapping between the memory state and the output variable.
A network receives a single input channel where pulses of varying width (100--230~ms) and sign are presented in sequence.
The network must report the XOR of the two input pulses by delivering an output pulse of appropriate sign after the second of the two input pulses.
A non-spiking \gls{rnn} ($\hat{N} = 64$) was trained to perform the temporal XOR task, using \gls{bptt} with \gls{mse} loss against the target output signal (target and output signals shown in Figure~\ref{fig:temporal_xor}a).
After 20 epochs of training with 500 samples per epoch, the \gls{rnn} reached negligible error on 200 test samples ($\approx 100\%$ accuracy).
A spiking ADS network ($N = 320$) was then trained to perform the task, reaching equivalent accuracy (Figure~\ref{fig:temporal_xor}a, b).

% Temporal XOR task figure

% Our method is able to learn multi-dimensional complex temporal supervised learning tasks
\subsection*{Wake-phrase detection}
\begin{figure}
  \includegraphics[width=\columnwidth]{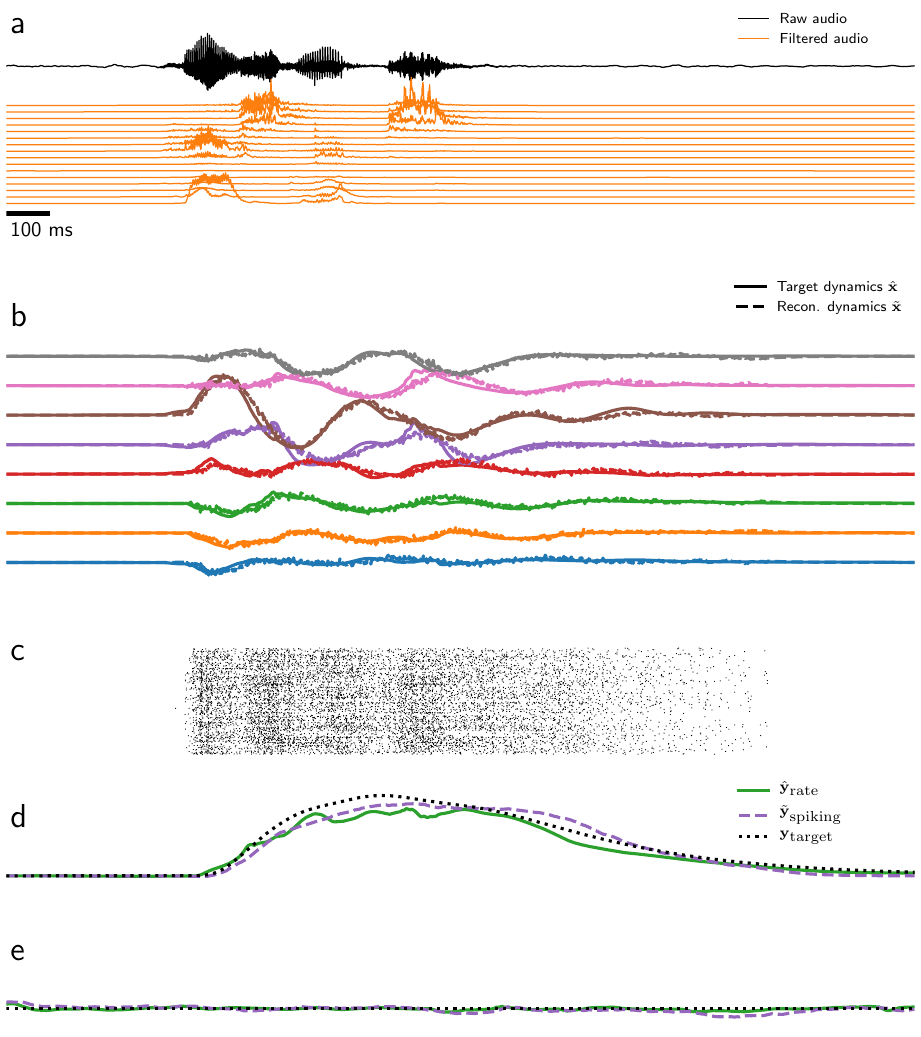}
  \centering
  \caption{
    \textbf{Our approach performs a supervised spoken audio multi-dimensional classification task with high accuracy.}
    \textbf{a} Audio samples were presented that either matched a spoken target phrase, or consisted of random speech or background noise.
    Raw audio (black) was filtered into 16 channels (orange) for classification by the network ($\textbf{c}\in\mathbb{R}^{16}$).
    \textbf{b} Internal \gls{rnn} dynamics ($\hat{\textbf{x}}$; solid) was reconstructed accurately by the trained spiking ADS network ($\tilde{\textbf{x}}$; dashed) based on the spiking activity (shown in \textbf{c}).
    An output signal was high (shown in \textbf{d}) when the audio sample matched the target signal, or low (shown in \textbf{e}) when the audio signal was background noise or other speech. Panels \textbf{a}--\textbf{d} correspond to a single trial.
  }
  \label{fig:audio_task}
\end{figure}

The temporal XOR task demonstrates that one-dimensional nonlinear tasks requiring memory can be learned through our method through supervised training.
To show that our approach also works on more realistic tasks with complex input dynamics, we implemented an audio wake-phrase detection task (Figure~\ref{fig:audio_task}; see Methods).
Briefly, real-time audio signals were extracted from a database of spoken wake phrases (``Hey Snips'' dataset\cite{snips}), or from a database of noise samples (``DEMAND'' dataset\cite{demand}).
The target wake phrase data was augmented with noise at an SNR of 10dB, then passed through a bank of 16 Butterworth filters with central frequencies spaced between 0.4 and 2.8 kHz (Figure~\ref{fig:audio_task}b).
We trained a non-spiking \gls{rnn} ($\hat{N}=128$) to perform the task with high accuracy, using \gls{bptt} under an \gls{mse} loss function against a smooth target classification signal (Figure~\ref{fig:audio_task}d, e).
We then trained a spiking ADS network ($N=768$) to implement the audio classification task.
The non-spiking \gls{rnn} achieved a testing accuracy of $\approx 90\%$, and our spiking imitator achieved $\approx 87\%$ after training for 10 epochs on 1000 training samples.

% Audio classification task figure

% How did we get our method to work?
\subsection*{Training considerations}
We found that slower input, internal and target dynamics in the \gls{rnn} were easier for the \gls{snn} to reconstruct than very rapid dynamics, depending on the neuron and synaptic time constants in the \gls{snn}. 
Longer and slower target responses yielded smoother ANN dynamics, which were easier for the spiking ADS network to learn.
Our approach did not assume any dendritic non-linearities, or multi-compartmental dendrites with complex basis functions.
Instead, the non-linearity of the spiking neuron dynamics is sufficient to learn the dynamics of a non-spiking ANN using the $\tanh$ nonlinearity.

We found that including a learning schedule for the error feedback rate~$k$ was important to achieve low reconstruction error.
The factor $k$ must drop to close to zero before the end of training, or else the \gls{snn} learns to rely on error feedback for accuracy, and generalisation will be poor once error feedback is removed.
Conversely, if $k$ drops too rapidly during training, the \gls{snn} is not held close to the desired target dynamics, and is unable to correctly learn the slow feedback weights $\Omega^{\textbf{s}}$.
For these reasons, a well-chosen schedule for $k$ is important during learning.
In this work we chose a progressive stepping function that decrements $k$ by a fixed amount after some number of signal iterations (see Methods).
Setting $k$ to a fixed value for some number of trials enables the \gls{snn} to adapt to the corresponding scale of error feedback by updating $\Omega^{\textbf{s}}$.

% Our method is robust to several sources of noise
\subsection*{Robustness to noise sources}
\begin{figure}
\centering
  \includegraphics[width=\columnwidth]{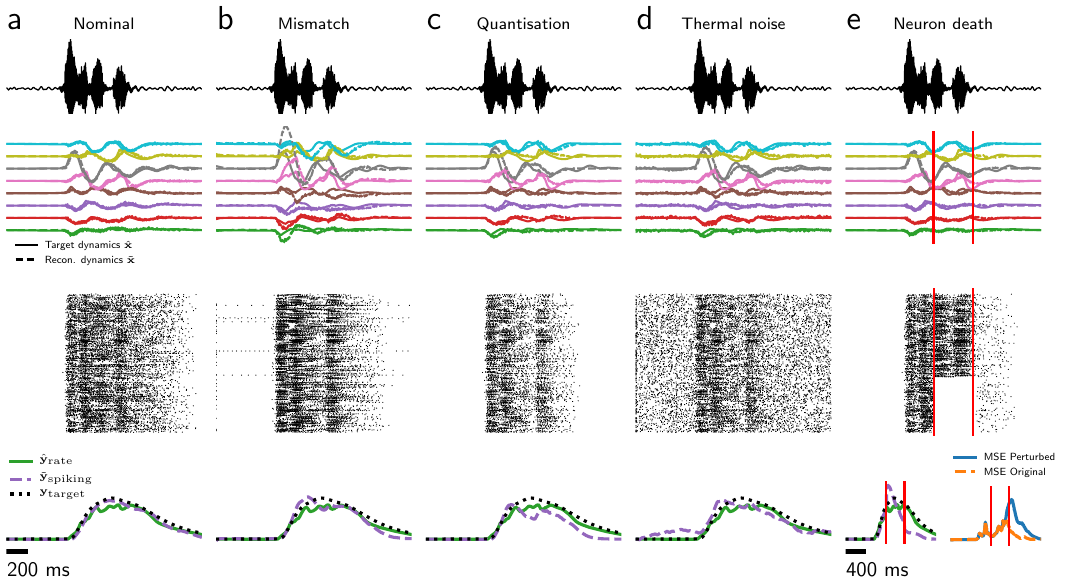}
  \caption{
    \textbf{Our trained spiking networks are robust to device mismatch and other sources of noise.}
    Each column shows (top to bottom) the raw signal input; the ANN and reconstructed dynamics; the spiking activity of the ADS network; and the task output and target signals.
    \textbf{a}~The trained ADS network reproduces the ANN internal and target signals with high accuracy.
    \textbf{b}~In the presence of simulated mismatch in a mixed-signal silicon implementation of LIF neurons (20\%; see Methods), the ADS network compensates well for the resulting frozen parameter noise.
    \textbf{c}~Frozen weight noise introduced by quantisation of weights to 4 bits is compensated by the ADS spiking network.
    \textbf{d}~The spiking network also compensates well in the presence of simulated thermal noise ($\sigma=5\%$).
    \textbf{e}~The balanced fast recurrent feedback connections $\Omega^{\textbf{f}}$ permit the ADS spiking network to compensate for sudden neuron death (40\% of spiking neurons silenced between vertical bars).
  }
  \label{fig:noise_sources}
\end{figure}

The slow learned recurrent feedback connections $\Omega^{\textbf{s}}$ in the spiking network enables the \gls{snn} to reproduce a learned task.
In contrast, the balanced fast recurrent feedback connections $\Omega^{\textbf{f}}$ are designed to enable the \gls{snn} to encode the dynamic variables $\tilde{\textbf{x}}$ in a way that is robust to perturbation\cite{boerlin_etal_2013, Deneve2016}.
We examined the robustness of our trained networks to several sources of noise (Figure~\ref{fig:noise_sources}).

\paragraph*{Device Mismatch}
% Robust to device mismatch
We first introduced frozen parameter noise as a simulation of device mismatch present in mixed-signal neuromorphic implementation of event-driven neuron and synapses.
We measured distributions of neuronal and synaptic parameters induced in silicon spiking neurons by device mismatch (see Methods; Figure~\ref{fig:mismatch_distribution}).
Measurements were performed on 1 core of 256 analog neurons and synapses, on fabricated mixed-signal neuromorphic DYNAP™-SE processors\cite{dynapse}.
We observed a consistent relationship between the mean and variance of parameter distributions: the variance of the measured parameters increased linearly with the magnitude of the set parameter.
We used this experimentally-recorded relationship to simulate mismatch in our spiking network implementations, simulating deployment of the networks on mixed-signal neuromorphic hardware.
Mismatched parameters $\Theta'$ were generated with $\Theta' \sim \mathcal{N}(\Theta, \delta\Theta)$, where $\delta$ determines the level of mismatch, which we found experimentally to be between 10--20\%.
Under 20\% simulated mismatch on weights, thresholds, biases, synaptic and neuronal time constants, our networks compensated well for the frozen parameter noise present in mixed-signal deployment (Figure~\ref{fig:noise_sources}b).

% Noise robustness in our spiking ADS network

\paragraph*{Quantisation Noise}
% Robust to quantisation error
In contrast to 64-bit floating point precision used by the non-spiking \gls{rnn}, deployment of NN architectures in memory-constrained systems often uses low bit-depth precision for weights and neuron state.
Mixed-signal neuromorphic architectures use analog voltages or currents to represent internal neural state, but can use some form of quantisation for synaptic weights.
For example, DYNAP™-SE2 processors impose a five-bit representation of synaptic weights, as well as a restricted fan-in of 64~pre-synaptic input sources per neuron\cite{dynapse}.
We imposed weight quantisation constraints on our spiking model, and found that our networks compensated well for the resulting frozen quantisation noise (Figure~\ref{fig:noise_sources}c; see Methods).

\paragraph*{Thermal Noise}
% Robust to thermal noise
Due to the analog representation of neuron and synapse states in mixed-signal neuromorphic chips, these state variables are subject to thermal noise.
Thermal noise appears as white-noise stochastic fluctuations of all states.
We simulated thermal noise by adding noise $\zeta \sim \mathcal{N}(0, \sigma)$ to membrane potentials $V$, with $\sigma = 1\%, 5\%, 10\%$ scaled to the range between reset and threshold potentials $V_\mathrm{reset}$ and $V_\mathrm{thresh}$.
The spiking ADS network performed well in the presence of thermal noise (Figure~\ref{fig:noise_sources}d).

\begin{figure}[h!]
  \includegraphics[width=\columnwidth]{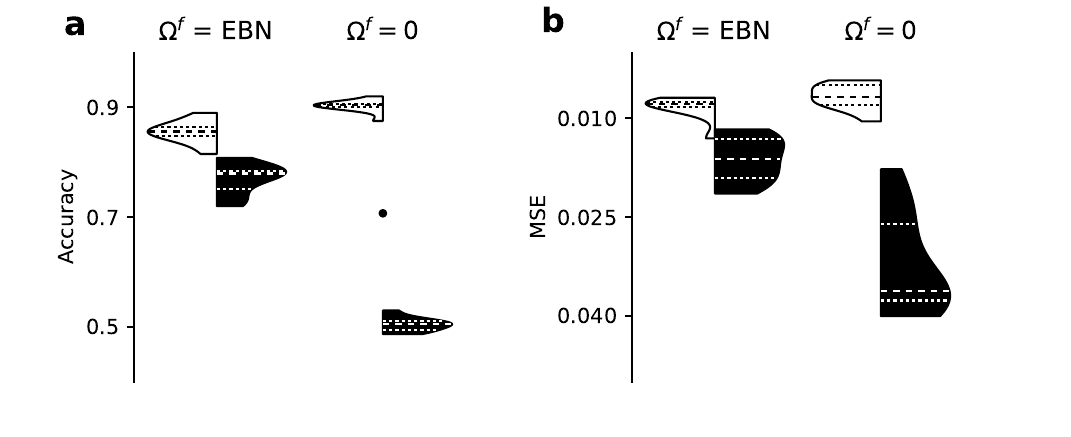}
  \centering
  \caption{
    \textbf{Fast balanced feedback $\Omega^{\textbf{f}}$ provides robustness to neuron silencing.}
    The effect of silencing 40\% of neurons in spiking ADS networks is shown on accuracy (\textbf{a}) and on output error \gls{mse} (\textbf{b}).
    White: Network performance without silencing.
    Black: Network performance when silencing 40\% of spiking neurons.
    Fast balanced recurrent feedback ($\Omega^{\textbf{f}} = \textrm{EBN}$) provided robustness to neuron silencing (small drop in performance).
    Networks without fast balanced recurrent feedback ($\Omega^{\textbf{f}} = 0$) exhibited more severe performance degradation under neuron silencing (large drop in performance).
  }
  \label{fig:neuron_failure}
\end{figure}

\paragraph*{Sudden Neuron Failure}
% Robust to neuron silencing
The fast recurrent feedback connections $\Omega^{\textbf{f}}$ present in spiking balanced networks have been shown to be able to compensate for neuron loss, where a subpopulation of spiking neurons is silenced during a trial\cite{boerlin_etal_2013, barrett_etal_2016, alemi_etal_2018_learning}.
We examined this property in our spiking ADS networks that include fast balanced feedback, and found that indeed our networks compensated well for neuron loss (Figure~\ref{fig:noise_sources}e).
In the absence of fast recurrent feedback (i.e. $\Omega^{\textbf{f}} = 0$), neuron silencing degraded the performance of the spiking ADS networks (Figure~\ref{fig:neuron_failure}).

% How does our approach compare to other standard approaches in robustness?
\subsection*{Comparison with alternative architectures}
We have demonstrated that our method produces spiking implementations of arbitrary tasks, defined through supervised training.
We compared our approach against several alternative methods for supervised training of \gls{snn}s, and evaluated the performance of these methods under simulated deployment on mixed-signal neuromorphic hardware:
\begin{itemize}[topsep=0pt, leftmargin=0pt]
    \item Reservoir Computing, in the form of a Liquid State Machine\cite{Maass_etal_2002_lsm}, relies on the random dynamics of an \gls{snn} to project an input over a high-dimensional temporal basis.
    A readout is then trained to map the random temporal basis to a specified target signal, using regularised linear regression.
    Since perturbation of the weights and neural parameters will directly modify the temporal basis, we expect the Reservoir approach to perform poorly in the presence of mismatch.
    \item The spiking FORCE algorithm\cite{Nicola2017} trains an \gls{snn} to mimic a teacher dynamical system.
    We applied this algorithm to a trained non-spiking \gls{rnn}s to produce a trained \gls{snn}, similarly as in our spiking ADS approach.
    \item We implemented the \gls{bptt} algorithm to train an \gls{snn} end-to-end, using a surrogate gradient function similar to ref.~\citenum{neftci_bp}.
    During training, these networks received input and target functions identical to those presented to the non-spiking \gls{rnn}.
\end{itemize}

\begin{figure}
\centering
  \includegraphics[width=\columnwidth]{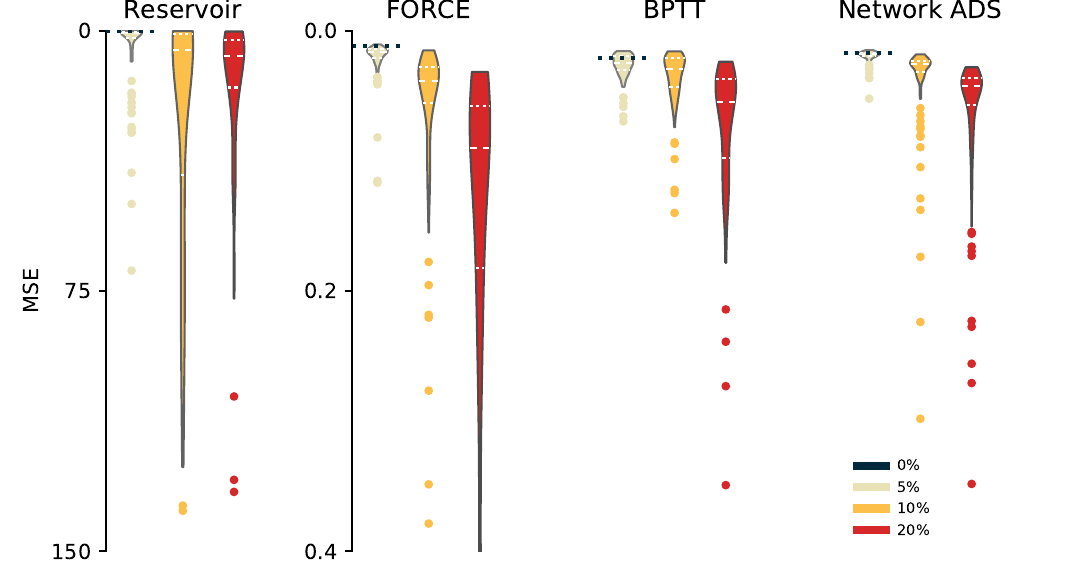}
  \caption{
    \textbf{Under simulated deployment, our method is more robust to mismatch than standard training approaches.}
    Shown are the distributions of errors (\gls{mse}) between the non-spiking teacher \gls{rnn} and final trained \gls{snn} response for 10 random initialisations of training for each architecture and 10 random instantiations for each level of mismatch.
    Dashed black line: baseline network with no mismatch ($\delta = 0\%$) for each architecture.
    See text for statistical comparisons.
  }
  \label{fig:mismatch_comparison}
\end{figure}

% Robustness to mismatch
We first examined simulated deployment of all architectures by simulating parameter mismatch (Figure~\ref{fig:mismatch_comparison}; see Methods).
We trained 10 networks for each architecture, and evaluated each network at three levels of mismatch ($\delta = 5\%, 10\%, 20\%$) for 10 random mismatch trials of 500 samples each.
We quantified the effect of mismatch on the performance of each network architecture by measuring the \gls{mse} between the \gls{snn}-generated output $\tilde{\textbf{y}}$ and the training target for that architecture.
For the FORCE and ADS networks the training target was the output of the non-spiking \gls{rnn}~$\textbf{y}$.
In the case of the Reservoir and BPTT architectures, the training target was the target task output $\hat{\textbf{y}}$.
Under the lowest level of simulated mismatch (5\%), the spiking ADS network showed the smallest degradation of network response (\gls{mse} drop 0.0094$\rightarrow$0.0109; $p \approx \num{8e-14}$, U test).
The spiking ADS network also showed the smallest mismatched variance in \gls{mse}, reflecting that all mismatched networks responses were close to the desired target response (\gls{mse} std. dev. ADS~0.0076; Reservoir~11.4; FORCE~0.0244; BPTT~0.0105; $p < \num{1e-2}$ in all cases, Levene test).
The spiking Reservoir architecture fared the worst, with large degradation in \gls{mse} for even 5\% mismatch (\gls{mse} drop 0.0157$\rightarrow$1.2523; $p \approx \num{4e-51}$, U test).
At 10\% simulated mismatch, comparable with deployment on mixed-signal neuromorphic devices, our spiking ADS network architecture maintained the best \gls{mse} (ADS~0.0161; Reservoir~6.19; FORCE~0.301; BPTT~0.308), performing significantly better than all other architectures ($p < \num{1e-6}$ in all cases, U test).
At 20\% simulated mismatch the performance of all architectures began to degrade, but our spiking ADS architecture maintained the best \gls{mse} (ADS~0.0470; Reservoir~10.5; FORCE~0.953; BPTT~0.565; $p < \num{5e-2}$ in all cases, U test).

% Mismatch robustness comparison figure

\begin{figure}[h!]
  \includegraphics[width=\columnwidth]{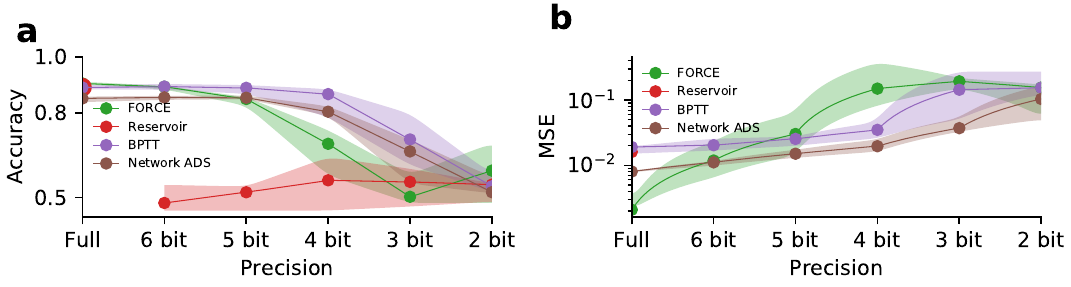}
  \centering
  \caption{
    \textbf{Our method is more robust to quantisation noise than other standard training approaches.}
    Median and \gls{iqr} for accuracy (\textbf{a}) and \gls{mse} (\textbf{b}) for four network architectures.
    The weights of 10 networks for each architecture were quantised to the bit-depths indicated, then evaluated on 1000 test samples each.
    Reservoir performance degraded completely for all quantisation bit-depths (very high \gls{mse}) so is not indicated in \textbf{b}.
  }
  \label{fig:quantisation}
\end{figure}

% Robustness to quantisation error
We compared the effect of quantisation noise on the four architectures, examining \numrange{6}{2} bits of weight precision (Figure~\ref{fig:quantisation}; see Methods).
Note that no architectures were trained using quantisation-aware methods, making this a direct test of inherent robustness to quantisation noise.
The Reservoir architecture broke down for any quantisation level (chance task performance accuracy $\approx \SI{50}{\percent}$).
The FORCE architecture performed well down to 5 bits (median accuracy 85\%), beyond which \gls{mse} increased and performance decayed to chance level at 3 bits  (med. acc. 50\%).
Both the ADS spiking network and BPTT architectures maintained good performance down to 4 bits of precision  (med. acc. ADS~81\%; BPTT~87\%), decaying to chance level at 2 bits (med. acc. ADS~52\%; BPTT~54\%).

\begin{figure}[h!]
    \centering
    \includegraphics[width=\columnwidth]{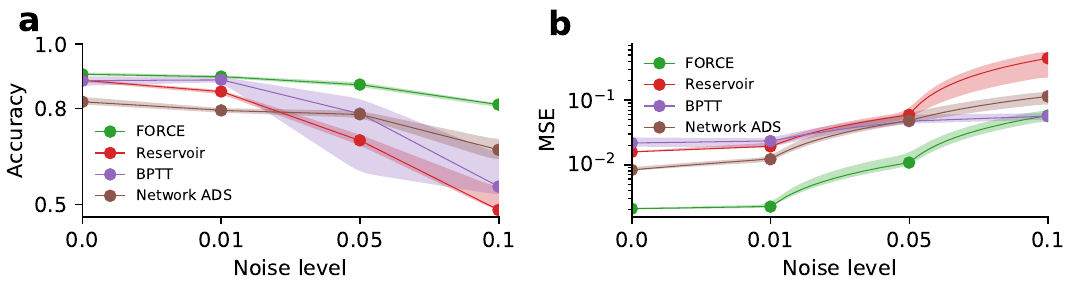}
    \caption{
        \textbf{Effect of membrane potential noise on network responses.}
        Median and \gls{iqr} for accuracy (\textbf{a}) and \gls{mse} (\textbf{b}) for four network architectures.
        Noise was added to 10 instances of each network architecture, as Normally-distributed noise with std. dev. $\sigma$ scaled to the range between $V_\textrm{thresh}$ and $V_\textrm{reset}$ for each neuron (see Methods).
    }
    \label{fig:membrane_potential_noise_comparison}
\end{figure}

% Robustness to thermal noise
We compared the effect of thermal noise of the four architectures, simulated as membrane potential noise (Figure~\ref{fig:membrane_potential_noise_comparison}; see Methods).
The FORCE architecture was most robust to thermal noise, performing best at all noise levels  (higher accuracy, $p < \num{5e-2}$; lower \gls{mse}, $p < \num{1e-3}$ except for highest level of noise; U tests).
All other architectures degraded progressively with increasing noise levels.
Our spiking ADS network architecture showed the smallest degradation in general over increasing noise levels  (\gls{mse} \numrange{0.0083}{0.115}; acc. \SIrange{82}{67}{\percent}).
The BPTT architecture also fared well, while dropping in accuracy for the largest noise level (med. acc. 56\%).

\subsection*{Power comparison for mixed-signal and traditional implementations}
We estimated and compared the power requirements between a direct implementation of the recurrent non-spiking network dynamics on commodity and ASIC hardware, against our mixed-signal spiking implementation of the network dynamics.
We performed the power comparison for the real-time audio processing task outlined above, for varying recurrent network dimensions.
Computation on the Dynap™SE1 processor occurs continuously in real-time, with no clock.
We selected the slowest clock speeds for the commodity hardware that are sufficient to support real-time operation.

We estimated the power requirements for an ultra-low-power digital microcontroller from ST Microelectronics (STM32L552xx)~\cite{stm32l552xx} (See Table~\ref{tab:power_comparison}).
When operating at a \SI{16}{\mega\hertz} clock frequency and efficiently implementing only the recurrent dynamics required by a $\hat{N}=64$-neuron non-spiking \gls{rnn}, the low-power MCU was estimated to require \SI{260}{\micro\watt} when simulating with a time-step $\textrm{dt} = \SI{10}{\milli\second}$, increasing to \SI{1130}{\micro\watt} for $\textrm{dt} = \SI{1}{\milli\second}$.
For the equivalent spiking network with $N=768$ spiking neurons the Dynap™-SE1 processor requires \SI{288}{\micro\watt} when fabricated at \SI{180}{\nano\meter} process, and \SI{38}{\micro\watt} when fabricated at \SI{65}{\nano\meter} process.
For larger non-spiking \glspl{rnn}, the Dynap™-SE1 processor has an increasing energy advantage over the low-power MCU.

We also considered the implementation of the non-spiking \gls{rnn} on an ultra-low-power ASIC, EIE~\cite{song_etal_2016}.
When implementing the dynamics required by a $\hat{N}=64$-neuron non-spiking \gls{rnn}, the ASIC required \SI{11}{\micro\watt} when simulating with a time-step $\textrm{dt}=\SI{10}{\milli\second}$, increasing to \SI{105}{\micro\watt} for $\textrm{dt} = \SI{1}{\milli\second}$.
The ASIC displays a power advantage when simulating dynamics for extremely small \glspl{rnn} with $\hat{N} < 35$, or with large time-steps $\textrm{dt}=\SI{10}{\milli\second}$ and $\hat{N} < 200$.
For larger networks and with more accurate temporal dynamics, the mixed-signal \gls{snn} implementation using our approach is more energy-efficient.
For further details of the power estimations see Methods and Table~\ref{tab:power_comparison}.

\section*{Discussion}
% Recap:
% What did we do?
% What did we show?
We propose a method for supervised training of spiking neural networks that can be deployed on mixed-signal neuromorphic hardware without requiring per-device retraining or calibration.
Our approach interprets the activity of a non-spiking \gls{rnn} as a teacher dynamical system.
Using results from dynamical systems learning theory, our spiking networks learn to copy the pre-trained \gls{rnn} and therefore perform arbitrary tasks over temporal signals.
Our method is able to produce spiking networks that perform both simple and complex non-linear temporal detection and classification tasks.
We show that our networks are considerably more robust to several forms of parameter and state noise, compared with several other common techniques for training spiking networks.

% What is the consequence of our work?
Our networks are by design robust to common sources of network and parameter variation, both intra- and inter-chip, which must be compensated for when deploying to mixed-signal neuromorphic hardware.
For levels of mismatch measured directly from neuromorphic devices, we show that common \gls{snn} network architectures break down badly.
Usual approaches for compensating for mismatch-induced parameter variation on neuromorphic hardware employ either on-device training \cite{Fusi2000, Cameron2008, Mitra2009, Pfeil2013, Wunderlich2019} or per-device calibration \cite{CostasSantos2007, neftci_indiveri, Neftci2011, Aamir2018, Wunderlich2019}, entailing considerable additional expense in hardware complexity or testing time.
In contrast, our method produces spiking networks that do not require calibration or retraining to maintain performance after deployment.
As a result, our approach provides a solution for cost-efficient deployment of event-driven neuromorphic hardware.

The coding scheme used by our spiking networks has been shown to promote sparse firing\cite{boerlin_etal_2013}.
For mixed-signal neuromorphic hardware, power consumption is directly related to the network firing rate.
Our method therefore produces networks that consume little power compared with alternative architectures that use firing-rate encoding or do not promote sparse activity\cite{Nicola2017, BPTT, Maass_etal_2002_lsm}.

% Limitations in filling the gap
Our approach to obtain high-performing \gls{snn}s is at heart a knowledge-transfer approach, relying on copying the dynamics of a highly-performing non-spiking \gls{rnn}.
This two-step approach is needed because the learning rule for our \gls{snn} requires a task to be defined in terms of a dynamical system, and is not able to learn the dynamics of an arbitrary input--output mapping (see Supplementary Methods).
Consequently, our spiking networks can only perform as well as the pre-trained non-spiking \gls{rnn}, and require multiple training steps to build a network for a new task.
Nevertheless, training non-spiking \gls{rnn}s is efficient when using automatic differentiation, just-in-time compilation and automatic batching\cite{jax2018github}, and can be performed rapidly on GPUs.
Our approach trades off between training time on commodity hardware, and immediate deployment on neuromorphic hardware with no per-device training required.

The robustness of our spiking ADS networks comes partially from the fast balanced recurrent feedback connections, which ensure sparse encoding and compensate in real-time for encoding errors\cite{boerlin_etal_2013, Deneve2016}.
These weights also degrade under noise, but can be adapted in a local untrained fashion using local learning rules that are compatible with HW implementation\cite{buchel}.

% Limitations in generalisation
Our supervised training approach is designed for temporal tasks, where input and target output signals evolve continuously.
This set of tasks encompasses real-time ML-based signal processing and recognition, but is a poorer fit to high-resolution frame-based tasks such as frame-based image processing.
These "one-shot" tasks can be mapped into the temporal domain by serialising input frames\cite{seq_mnist} or by using temporal coding schemes\cite{Thorpe_etal2001}.
We found anecdotally that temporal discontinuities in input and target time series made training our ADS networks more difficult, with the implication that a careful matching between task and network time constants is important.

Our approach builds single-population recurrent spiking networks, in contrast to deep non-recurrent network architectures which are common in 2021\cite{dl_overview}.
Recurrent spiking networks such as \glspl{lsm} have been shown to be universal function approximators\cite{lsm_universal}, but \glspl{rnn} do not perform the progressive task decomposition that can appear in deep feed-forward networks\cite{interpretability}.
Interpretability of the internal state of recurrent networks such as ours is therefore potentially more difficult than for deep feedforward architectures.

% How did we move the field forward?
Neuromorphic implementation of spiking neural networks has been hailed as the next generation of computing technology, with the potential to bring ultra-low-power non-von-Neumann computation to embedded devices.
However, parameter mismatch has been a severe hurdle to large-scale deployment of mixed-signal neuromorphic hardware, as it directly attacks the reliability of the computational elements --- a problem that commodity digital hardware generally does not face.
Previous solutions to device mismatch have been impractical, as they require expensive per-device calibration or training prior to deployment, or increased hardware complexity (and therefore cost) in the form of on-device learning circuits.
We have provided a programming method for mixed-signal neuromorphic hardware that frees application developers from the necessity to worry about computational unreliability, and does not require per-device handling during or after deployment.
Our approach therefore removes a significant obstacle to the large-scale and low-cost deployment of neuromorphic devices.

\section*{Methods}
% 3000 words
% Current count: 1000 words

We trained and simulated ANNs and \glspl{snn} using Rockpool\cite{rockpool}, an open-source Python package for machine learning of \glspl{snn}.
We implemented a liquid state machine \gls{snn}\cite{Maass_etal_2002_lsm}; spiking FORCE network\cite{Nicola2017}; and a \gls{bptt}-trained \gls{snn}\cite{BPTT} using Jax\cite{jax2018github} and custom-written forward-Euler solvers.
Parameters for all architectures are given in the Supplementary Material.
Code to generate all models, analysis and figures in this paper are available from \url{https://github.com/synsense/Robust-Classification-EBN}.

\subsection*{Temporal XOR task}
We created signals of a total duration of 1 second, of which the first two thirds were dedicated to the input and the last third to the target (Figure~\ref{fig:temporal_xor}).
During the input time-frame, two activity bumps were created on a single input channel representing the binary inputs to the logical XOR operation.
The bumps had varying length (uniformly drawn between \SIrange{66}{157}{\milli\second}) and magnitude $\pm1$, and were smoothed with a Gaussian filter to produce smooth activity transitions.
In the final third of the signal we defined a target bump of magnitude $\pm1$, indicating the true output of the XOR operation.
The target bump was also smoothed with a Gaussian filter.
We trained a rate network ($\hat{N}=64$) to high performance on the XOR task, then subsequently trained a spiking model ($N=320$) to follow the dynamics of the trained rate network .
We used a fixed learning rate $\eta = \num{1e-5}$ and fixed error feedback rate $k = 75$ during \gls{snn} training.
Output classification from both networks was determined by the network output passing the thresholds $\pm0.5$.

\subsection*{Speech classification task}
We drew samples from the ``Hey Snips'' dataset\cite{snips}, augmented with noise samples from the DEMAND dataset\cite{demand}, with a signal-to-noise ratio of \SI{10}{\deci\bel}.
Each signal had a fixed length of \SI{5}{\second} and was pre-processed using a 16-channel bank of 2nd-order Butterworth filters with evenly-spaced centre frequencies ranging \SIrange{0.4}{2.8}{\kilo\hertz}.
The output of each filter was rectified with $\mathrm{abs}(\cdot)$, then smoothed with a 2nd order Butterworth low-pass filter with cut-off frequency \SI{0.3}{\kilo\hertz} to provide an estimate of the instantaneous power in each frequency band.
The rate network for the speech classification task ($\hat{N}=128$) was trained for 1 epoch on \num{10000} samples to achieve roughly the same performance as the spiking network trained with \gls{bptt}.
We trained spiking networks ($N=768$; $\tau_{\mathrm{mem}}= \SI{50}{\milli\second}$; $\tau_{\mathrm{fast}} = \SI{1}{\milli\second}$; $\tau_{\mathrm{slow}} = \SI{70}{\milli\second}$) for 5 epochs on 1000 training samples, validated on 500 validation samples and 1000 test samples.
To perform a classification we integrated the output of the network when it passed a threshold of 0.5.
We then applied a subsequent threshold on this integral, determined by a validation set, to determine the final prediction.
We used a fixed learning rate $\eta = \num{1e-4}$ and a decaying step function for the error feedback factor $k$ (from \numrange{200}{25} in 8 evenly-spaced steps).

\subsection*{Spiking neuron model and initialisation}
We used an \gls{lif} neuron model with a membrane time constant $\tau_\textrm{mem} = \SI{50}{\milli\second}$; reset potential $V_\textrm{reset} = 0$; resting potential $V_\textrm{rest} = 0.5$; and spiking threshold $V_\textrm{thresh}= 1$.
The membrane potential dynamics for the neuron model were given by
\begin{equation*}
    \tau_{\mathrm{mem}} \frac{\partial V}{\partial t} = V_{\mathrm{rest}}-V + I_{\mathrm{inp}} +  I_{\mathrm{fast}} + I_{\mathrm{slow}} + I_{\mathbf{e}} + \eta_n
\end{equation*}
with input current $I_{\mathrm{inp}}$; fast and slow recurrent post-synaptic potentials (PSPs) $I_{\mathrm{fast}}$ and $I_{\mathrm{slow}}$; error current $I_{\mathbf{e}}  = k\mathbf{D}^T\mathbf{e}$; and noise current $\eta$.
Output spikes from a neuron are given by $o(t) = V > V_\textrm{thresh}$.
Synaptic dynamics were described by
\begin{equation*}
    \tau_{\mathrm{syn}} \frac{\partial I_*}{\partial t} = -I_* + (W \mathbf{o}(t) \tau_{\mathrm{syn}}) / {\Delta t}
\end{equation*}
with input synaptic weights $W$; synaptic time constants $\tau_{\mathrm{syn}} = \SI{1}{\milli\second}$ and $\SI{70}{\milli\second}$ for fast and slow synapses, respectively; and simulation time step~$\Delta t$.
Feed-forward and decoding weights were initialised using a standard normal distribution scaled by the number of input/output dimensions ($\hat{N}$).
Fast balanced recurrent feedback connections were initialised and rescaled according to the threshold and reset potential, as described in ref. \citenum{boerlin_etal_2013}. The spiking network was simulated using a forward Euler solver with a simulation time step of \SI{1}{\milli\second}.

\subsection*{Non-spiking network}
The dynamics of a neuron in the non-spiking \gls{rnn} were described by
\begin{equation*}
\tau_j \dot{x}_j = -x_j + \hat{\mathbf{F}}c_j(t) + \hat{\Omega}f(\textbf{x}) + b_j + \epsilon_j    
\end{equation*}
with input $c(t)$; encoding weights $\hat{\textbf{F}}$; recurrent weights $\hat{\Omega}$; non-linearity $f(\cdot) = \textrm{tanh}(\cdot)$; bias $b$; and noise term $\epsilon$.
Time constants $\tau$ were initialised with linearly spaced values (\SIrange{10}{100}{\milli\second}).
The trainable parameters in this network are the time constants $\tau$; the encoding and recurrent weights $\hat{\mathbf{F}}$ and $\hat{\Omega}$; and the biases $b$.
No noise was applied during training or inference ($\epsilon = 0$).

\subsection*{Measurements of parameter mismatch}
Using recordings from fabricated mixed-signal neuromorphic chips we measured levels of parameter mismatch (i.e. fixed substrate noise pattern) present in hardware. In particular, for DYNAP™-SE\cite{dynapse}, a neuromorphic processor which emulates LIF neuron, AMPA and NMDA synapse models with analog circuits, we measured neuron and synaptic time constants, and synaptic weights for individual neuron units, by recording and analysing the voltage traces produced by these circuits. We observed levels of mismatch in the order of 10--20\% for individual parameters, with widths of the distributions being proportional to the means (see Figure~\ref{fig:mismatch_distribution}).

\subsection*{Power estimates}
Since the input and output weighting differs between the spiking and non-spiking network, and comprises only a small portion of the parameters, we limited our power comparison to the recurrent portion of the network.
Updating the recurrent dynamics for the non-spiking rate network requires multiply-accumulate operations for the recurrent input $\textbf{r}_t = \hat{\Omega}f(\textbf{x}_t)$ (neglecting the transfer function $f(\cdot)$); multiply-accumulate operations for the Euler solver update $\textbf{x}_{t+1} = \textbf{x}_t + \dot{\textbf{x}}_t * \textrm{d}t / \tau$; and accumulate operations for $\dot{\textbf{x}}_t = -\textbf{x}_t + \textbf{i}_t + \textbf{b} + \textrm{r}_t$.
With $\hat{N} = \num{64}$ neurons, these amount to 8576 OPs, with MACs counted as two OPs.
With a time-step of $\textrm{d}t = \SI{1}{\milli\second}$, this corresponds to $\SI{8.58}{GOPS}$ (Giga-OPs per second).
We estimated the power to implement our RNN on non-neuromorphic NN accelerators by using previously reported power as \SI{}{GOPS\per\watt}.
We examined only chips with published data for total power, and where we could identify the fabrication node for the published chip.
We re-scaled power estimates to normalise against the fabrication node, providing estimates for \SI{65}{\nano\meter} nodes in all cases.
For the ultra-low-power microcontroller (STM32L552xx), we assumed that the MCU switched to a low-power sleep mode once the dynamics for a given time-step were computed.
This permits the MCU to save power when only a portion of computing resources is required to simulate real-time dynamics.

Again neglecting synaptic operations required for input and output, we estimate the energy for routing a single recurrent spike on the DYNAP™-SE1 mixed-signal neuromorphic processor as \SI{3.3}{\nano\joule}.
We found that the firing rate of the spiking population is upper-bounded by approximately $\SI{3}{\hertz}$ per neuron during simulation.
For the spiking recurrent population with $N = \num{768}$, this corresponds to energy usage of $\SI{7.6}{\micro\watt}$ dynamic power consumption.
Static power consumption for the DYNAP™-SE1 processor is estimated at $\SI{30}{\micro\watt}$.
Table \ref{tab:power_comparison} compares the energy consumption of running the ANN on an efficient ASIC~\cite{EIE}, and a low-power general purpose MCU~\cite{st} to the energy consumption of the DYNAP™-SE1 using the spiking network with 12 times more neurons.

\subsection*{Simulated mismatch}
To simulate parameter mismatch in mixed-signal neuromorphic hardware we derived a model where the values for each parameter follow a normal distribution with the standard deviation depending linearly on the mean. The mismatched parameters $\Theta'$ are obtained with $\Theta' \sim \mathcal{N}(\Theta, \delta\Theta)$ where $\delta$ determines the level of mismatch. We considered three levels of mismatch: 5, 10 and 20\%.

\subsection*{Quantisation noise}
We introduced quantisation noise by reducing the bit-precision of all weights post-training to 2,3,4,5 and 6 bits. The weights were quantised by setting $\mathbf{W_{\mathrm{disc}}^s} = \rho \nint{W / \rho}$ where $\rho = (\mathrm{max} (\mathbf{W_{\mathrm{full}}^s})- \mathrm{min} (\mathbf{W_{\mathrm{full}}^s})) / (2^b-1)$ and $\nint{.}$ is the rounding operator.

\subsection*{Simulated thermal noise}
Thermal noise is inherent in neuromorphic devices and can be modeled by Gaussian noise on the input currents.
We applied three different levels of thermal noise ($\sigma=0.01, 0.05, 0.1$) that was scaled according to the difference between $V_\mathrm{reset}$ and $V_\mathrm{thresh}$ to assure equal amounts of noise for neuron model and network architecture.

\subsection*{Neuron silencing}
We created four network instances grouped into two pairs: One pair was trained with the fast recurrent feedback connections $\Omega^{\textbf{f}}$ as described above, and the other pair with $\Omega^{\textbf{f}} = 0$.
We then clamped 40\% of the neurons of one instance of both pairs to $V_{\mathrm{reset}}$ while evaluating 1000 test samples.

\subsection*{Benchmark network architectures}

\subsection*{Statistical tests}
All statistical comparisons were double-sided Mann-Whitney U tests unless stated otherwise.

\section*{Acknowledgements}
This project has received funding in part by the European Union's Horizon 2020 ERC project NeuroAgents (Grant No. 724295); from the European Union's Horizon 2020 research and innovation programme for ECSEL grants ANDANTE (grant agreement No. 876925), TEMPO (grant agreement No. 826655), and SYNCH (grant agreement No. 824162); and from "Fondo di Ateneo per la ricerca 2020" (FAR2020) of the University of Sassari (grant to S. Solinas).

\section*{Author contributions}

DRM and GI conceived and designed the research.
JB and DRM developed software and simulations.
JB, SS and DZ performed experiments and collected data.
JB and DRM analysed and interpreted the data.
JB and DRM drafted the manuscript.
DRM, JB, DZ, SS and GI performed critical revision of the manuscript.
DRM approved the final version for publication.

\section*{Additional information}

The authors declare no competing interests.

Supplementary Information is available for this paper.

Correspondence and requests for materials should be addressed to Dylan R Muir (\url{dylan.muir@synsense.ai}) and Julian Buchel (\url{jubueche@ethz.ch}).

\section*{Data availability}
Code to generate all models, analysis and figures in this paper are available from \url{https://github.com/synsense/Robust-Classification-EBN}.

% -- Bibliography
% Target: 50 references (excluding methods-only refs)
% Current: 59 references total
\bibliographystyle{naturemag}
\bibliography{bibliography}

% Supplementary material
\newpage
\section*{Supplementary Material}
\beginsupplement

\begin{figure}[h!]
    \centering
    \includegraphics[width=\columnwidth]{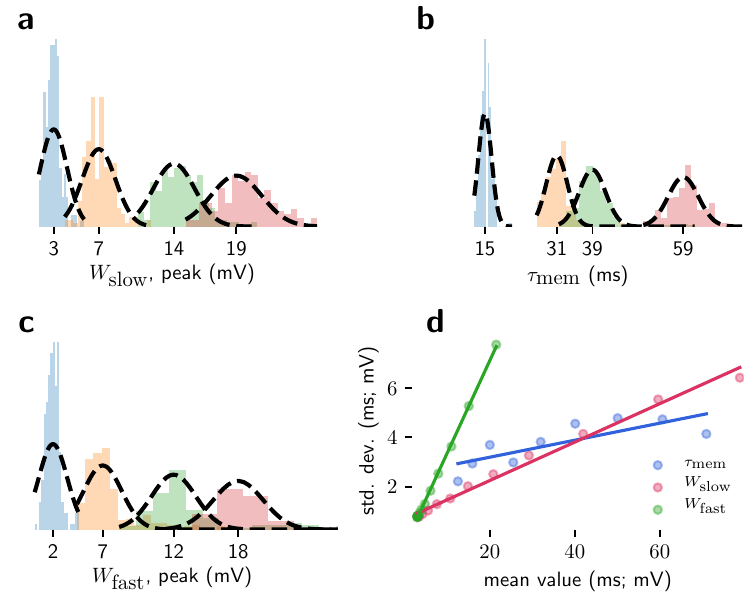}
    \caption{
        \textbf{Neuronal and synaptic parameter mismatch on mixed-signal Neuromorphic devices follow a mean-to-variance linear scaling rule.}
        Measurements of actual weights of two types of silicon synapses with slow (\textbf{a}) and fast dynamics (\textbf{c}), as well as neuron time constants (\textbf{b}), show a consistent linear relationship between nominal set value and the distribution of actual mismatch parameter values.
        In all cases, the variance of measured parameters scaled with the nominal set value (\textbf{d}) (linear regression $r= \num{0.79}$ $\tau_\textrm{mem}$; $r = \num{0.994}$ $W_\textrm{slow}$; $r = \num{0.9996}$ $W_\textrm{fast}$).
    }
    \label{fig:mismatch_distribution}
\end{figure}

\newpage

\newpage

\newpage

\newpage
\begin{table}[h]
\centering
\caption{
    Power estimations for a non-spiking \gls{rnn} and the corresponding \gls{snn} implemented on DYNAP™-SE1.
}
\resizebox{\columnwidth}{!}{%
\begin{tabular}{r|p{20ex}p{20ex}p{20ex}p{20ex}}
{ANN neurons $\hat{N}$} & STM32L552xx~\cite{stm32l552xx} @\SI{16}{\mega\hertz} (\SI{}{\milli\watt}) & STM32L552xx~\cite{stm32l552xx} @\SI{80}{\mega\hertz} (\SI{}{\milli\watt}) & EIE~\cite{song_etal_2016} \newline @\SI{65}{\nano\meter} (\SI{}{\milli\watt}) & Dynap™-SE1 @\SI{65}{\nano\meter} (\SI{}{\milli\watt}) \\ \midrule
32  & 0.4           & 0.45          & \textbf{0.027}   & 0.032   \\
64  & 1.13          & 1.31          & 0.11   & \textbf{0.038}   \\
128 & N/A$^\dagger$  & 4.8           & 0.41   & \textbf{0.057}   \\
256 & N/A$^\dagger$  & 11.4          & 1.6   & \textbf{0.13}   \\
512 & N/A$^\dagger$  & N/A$^\dagger$  & 6.5   & \textbf{0.43}
\end{tabular}%
}
$^\dagger$\small{For these parameter combinations, the \gls{rnn} required more computation than possible on the MCU for real-time operation.}
\label{tab:power_comparison}
\end{table}

\newpage
\section*{Supplementary Methods}
\subsection*{Learning in adaptive non-linear control theory} \label{sec:learning_adap}
Let us assume an arbitrary dynamical system of the form
\begin{equation} \label{eq:teacher_dyn}
  \dot{\textbf{x}}(t) = f(\textbf{x}(t)) + \textbf{c}(t)
\end{equation}
where $\textbf{x}(t)$ is a vector of state variables $x_j(t)$, $f(.)$ is a non-linear function (e.g. $\mathrm{tanh}(.)$), and $\textbf{c}(t)$ is a time-dependent input of the same dimensionality of $\textbf{x}(t)$.
Furthermore, let us assume a "student" dynamical system of the form
\begin{equation} \label{eq:student_dyn}
  \dot{\hat{\textbf{x}}}(t) = -\lambda \hat{\textbf{x}}(t) + \mathbf{W}^T \Psi(\hat{\textbf{x}}(t)) + \textbf{c}(t) + k \cdot \textbf{e}(t)
\end{equation}
where $\hat{\textbf{x}}(t)$ is a vector of state variables $\hat{x}_j(t)$, $\lambda$ is a leak term, and $\textbf{c}(t)$ is the same time-dependent input as in Equation \ref{eq:teacher_dyn}.
Over time, the signed error $\textbf{e}(t) = \textbf{x}(t) - \hat{\textbf{x}}(t)$ between the teacher dynamics (Eq. \ref{eq:teacher_dyn}) and the student dynamics (Eq. \ref{eq:student_dyn}) is computed and fed into the student dynamics, causing the student state variables $\hat{\textbf{x}}(t)$ to closely follow the target variables $\textbf{x}(t)$.

This close tracking enables us to update the weights $\mathbf{W}$ using Eq. \ref{eq:learning_rule}, so that over the course of learning, the factor $k$ can be reduced to zero and the network follows the teacher dynamics autonomously, using only a weighted sum of basis functions, given by $\Psi(\hat{\textbf{x}}(t))=\phi(\mathbf{M}\hat{\textbf{x}}(t)+\theta)$, for some non-linear function $\phi$, and some random $\mathbf{M}$ and $\theta$.

The learning rule used to adapt the weights $\mathbf{W}$ is given by
\begin{equation} \label{eq:learning_rule}
  \dot{\mathbf{W}} = \eta \Psi(\hat{\textbf{x}}(t))\textbf{e}(t)^T
\end{equation}
and can be shown to let the weights $\mathbf{W}$ converge towards the optimal weights, denoted $\mathbf{W}^\mathrm{true}$, assuming that the input $\textbf{c}(t)$ does not lie on a low-dimensional manifold and that the student system has enough high-dimensional basis functions. For more information and a proof of the above statement, see ref. \citenum{alemi_etal_2018_learning}

It should be noted that the relation between $x(t)$ and $c(t)$ should be well-defined by an autonomous (non-)linear dynamical system of the form $\dot{\textbf{x}}(t) = f(\textbf{x}(t))+\textbf{c}(t)$ and that one cannot simply assume a "black box" dynamical system implementing \textit{any} relation of the form $\dot{\textbf{x}}(t) = \mathcal{B}(\textbf{x}(t),\textbf{x}(t))$.
Concretely, in the light of classification, one cannot simply assume there exists an autonomous dynamical system relating the input~$\textbf{c}(t)$ to some target response variable, and that this relation can be learned by the learning rule described above.
This observation is important, as it makes it harder to build a classifier given the tools described above. In the next section, we will review how a network of spiking neurons can implement the above learning rule in order to learn the dynamics of a teacher dynamical system.

\subsection*{Learning arbitrary dynamical systems in \glspl{ebn}} \label{sec:learning_rule}
In this section, we will briefly recapitulate how an \gls{ebn} of spiking neurons can learn to implement any non-linear dynamical system of the form $\dot{\textbf{x}}(t)=f(\textbf{x})+\textbf{c}(t)$.
We will assume that, given a network of $N$ neurons, one can use a decoder $\mathbf{D}$ to reconstruct the target variable from the filtered spike trains of the population using $\hat{\textbf{x}}(t) = \mathbf{D}\textbf{r}(t)$, so that $\textbf{x}(t) \approx \hat{\textbf{x}}(t)$.

Derived from the fact that in an \gls{ebn} a neuron only fires a spike if it contributes to reducing the loss $L$ \cite{bourdoukan_etal_2012, boerlin_etal_2013}, given by
\begin{equation*}
  L = \frac{1}{T} \sum_{t=0}^T\|x(t)-\hat{x}(t)\|_2^2 + \mu\|r(t)\|_2^2 + \nu\|r(t)\|_1
\end{equation*}
with smoothed firing rates $r(t)$ and loss regularisation terms $\mu$ and $\nu$, the membrane potentials in the network are then given by
\begin{equation} \label{eq:voltage_dyn}
  V(t) = \mathbf{D}^T \mathbf{x}(t) - \mathbf{D}^T \mathbf{D}\mathbf{r}(t) - \mu \mathbf{r}(t)
\end{equation}

Following ref.~\citenum{alemi_etal_2018_learning}, we differentiate Eq.~\ref{eq:voltage_dyn} and substitute the smoothed firing rates $\dot{\mathbf{r}}(t)=-\lambda \mathbf{r}(t) + \mathbf{o}(t)$; the teacher dynamics $\dot{\mathbf{x}}(t)=f(\mathbf{x}(t))+\mathbf{c}(t)$; and the decoded dynamics of the student $\hat{\mathbf{x}}(t)=\mathbf{D}\mathbf{r}(t)$, to obtain
\begin{equation*}
\resizebox{1.0\hsize}{!}{
\begin{math}
  \begin{aligned}
    \dot{V}(t) &= -\lambda V(t) + \mathbf{D}^T(f(\mathbf{x}(t)+\mathbf{c}(t))) - \mathbf{D}^T\mathbf{D}(-\lambda \mathbf{r}(t)+\mathbf{o}(t)) - \mu (-\lambda \mathbf{r}(t) + \mathbf{o}(t)) \\
    &= - \lambda V(t) + \mathbf{D}^T \mathbf{c}(t) - (\mathbf{D}^T\mathbf{D}+\mu \textbf{I})\mathbf{o}(t) + \mathbf{D}^T(\lambda \mathbf{x}(t) + f(\mathbf{x}(t))) \\
  \end{aligned}
  \end{math}}
\end{equation*}
with decay rates $\lambda$; population spike trains $\textbf{o} = V > V_\textrm{thresh}$; and identity matrix~$\mathbf{I}$.

Under the non-linear control learning theoretical result of ref.~\citenum{alemi_etal_2018_learning}, the term $\lambda \mathbf{x}(t) + f(\mathbf{x}(t))$ is approximated by a weighted set of basis functions over slow recurrent feedback weights $\Omega^{\mathbf{s}}$, given by $\Omega^{\mathbf{s}} \Psi(\mathbf{r}(t))$, where $\Psi(\mathbf{r}(t))=\phi(\mathbf{M}\mathbf{r}(t)+\theta)$, inspired by complex multi-compartmental dendritic dynamics.
In our networks we omitted the term $\Psi(\mathbf{r}(t))$ and simply replaced it with the population spike train  $\mathbf{o}(t)$.

By feeding the error term $\mathbf{e}(t) = \mathbf{x}(t) - \hat{\mathbf{x}}(t)$ back into the network using the encoding/decoding weights $\mathbf{D}^T$, we obtain the final network dynamics
\begin{equation*}
  \dot{V}(t) = -\lambda V(t) + \mathbf{F}\mathbf{c}(t) - \Omega^{\mathbf{f}}\mathbf{o}(t) + \Omega^{\mathbf{s}}\mathbf{o}(t) + k\mathbf{D}^T\mathbf{e}(t)
\end{equation*}
where the encoding weights $\mathbf{F}$ are given by $\mathbf{D}^T$ and the optimal fast recurrent weights by $\Omega^{\mathbf{f}}=-(\mathbf{D}^T\mathbf{D} + \mu\mathbf{I})$\cite{boerlin_etal_2013}.

Similar to the case in control-theory, the learning rule for the slow recurrent weights is given by
\begin{equation*}
  \dot{\mathbf{\Omega}}^\mathbf{s} = \eta \Psi(\textbf{r}(t))(\mathbf{D}^T\textbf{e}(t))^T
\end{equation*}
where we simply replaced $\Psi(\textbf{r}(t))$ with $\textbf{r}(t)$.

To speed up the simulations, one could assume that the accumulated updates to $\Omega^{\mathbf{s}}$, given by $\sum_{t=0}^{T}{\eta \textbf{r}(t)(\mathbf{D}^T\textbf{e}(t))^T}$ are approximately the same as accumulating the rates and errors into large arrays of size $N \times T$ and performing the update in a batched fashion, after a whole signal was evaluated rather than after every time-step, using
\begin{equation*}
  \dot{\Omega}^\mathbf{s} = \eta \mathbf{r}(\mathbf{D}^T\mathbf{e})^T
\end{equation*}
We verified experimentally that this method is indeed faster, but yields less optimal results.

\subsection*{Spiking ADS network connectivity}
The decoder weights, denoted $\mathbf{D}$, are initialised using a standard normal distribution $~\mathcal{N}(0, 1/N_\textbf{c})$, where $N_\textbf{c}$ is the dimensionality of the input signal. The encoding weights $\textbf{F}$ are given by simply transposing the decoding weights such that $\textbf{F} = \textbf{D}^T$.

Unless stated otherwise, the slow recurrent connections were initialized using a zero matrix: $\Omega^{\mathbf{s}} = \mathbf{0}$.

Following the optimal network connectivity of \glspl{ebn}\cite{boerlin_etal_2013}, we initialized
$\Omega^{\mathbf{f_{\mathrm{*}}}}$ with $\mathbf{D^TD} + \mu \lambda_d^2 \mathbf{I}$, which was then transformed to $\Omega^{\mathbf{f}}$ using
\begin{equation*}
    \Omega^{\mathbf{f}} = \frac{\xi a}{\tau_{\mathrm{fast}}} \Omega^{\mathbf{f_{\mathrm{*}}}} / V^{*}_{\mathrm{thresh}}
\end{equation*}
where
\begin{equation*}
    V^{*}_{\mathrm{thresh,n}} = \frac{\nu \lambda_d + \mu \lambda_d^2 + \|D_n\|_2^2}{2}
\end{equation*}
with
$\mu = 0.0005$, $\nu = 0.0001$, $\lambda_d = 20$, $\xi = 10$, and $a = 0.5$.

We applied the transformation described in the section ``Scaling and physical units'' in ref. \cite{boerlin_etal_2013} to transform our network to have $V_{\mathrm{reset}}=0$ and $V_{\mathrm{thresh}} = 1.0$.

\subsection*{Training an \gls{ebn}-based classifier}
The coding properties of \glspl{ebn} make them attractive for many applications, especially in the neuromorphic community.
However, so far it has only been shown how to implement linear and non-linear dynamical systems of a specific form using \glspl{ebn}. In this section we provide a method to train an \gls{ebn} to perform classification tasks of varying
complexity.

Let us define our time-varying, real-valued input as $\textbf{c}(t)$, where $\textbf{c} \in \mathcal{R}^{N_c \times 1}$ and $t \in \{0...T\}$.
The goal of training a classifier is to find a mapping $f(\mathbf{c},\Theta) \rightarrow \mathbf{y}$ that maps any input $\mathbf{c}$ to the desired target variable $\mathbf{y}$, where $\mathbf{y}$ is a variable over time indicating the target and $\Theta$ is the set of system parameters.
For simplicity, we will consider the case of binary classification. We however note that our method can be easily extended to a multi-class classification task.

Considering the ability of the learning rule presented in section \ref{sec:learning_rule}, one might be be inclined to assume a "black box" dynamical system $\mathcal{B}$ of the form $\dot{\textbf{y}}(t) = f(\textbf{y}(t)) + \textbf{c}(t)$ that, given input $\mathbf{c}$, autonomously produces the desired target $\mathbf{y}$.
Two problems come with this approach:
\begin{enumerate}
    \item The dynamical system is not well-defined, as it is no-longer explicitly defined by a teacher dynamical system.
    Furthermore, the system is autonomous, as the function $f(\cdot)$ does not depend on the input, but only on past values of the target variable, making it impossible for the system to find a complex relationship between input and target.
    \item This approach assumes that the input and target variable have the same number of dimensions, which is almost never the case.
\end{enumerate}

How can we use the above learning rule to train an \gls{ebn} to perform arbitrary classification tasks at low metabolic cost, high robustness and good classification performance? To answer this question, we consider a simple \gls{rnn} comprising $\hat{N}$ units, following
\begin{equation} \label{eq:rate_dyn}
  \tau_j \dot{x}_j(t) = -x_j(t) + \hat{\mathbf{F}}c(t)_j + \hat{\mathbf{\Omega}}f(x(t))_j + b_j + \epsilon_j
\end{equation}
where $\tau_j$ is the time constant of the \textit{j}-th unit; $\hat{\mathbf{F}}$ are feed-forward encoding weights of shape $\hat{N} \times N_c$; $c(t)$ is the $N_c$ dimensional input at time $t$; $f(.)$ is a non-linear function (e.g. $\mathrm{tanh}(.)$); $\hat{\Omega}$ are recurrent weights of shape $\hat{N} \times \hat{N}$; and $b_j$ and $\epsilon_j$ are bias and noise terms, respectively.

We observe that an \gls{rnn} can be rewritten in the general form
\begin{IEEEeqnarray*}{rCl}
    \dot{\mathbf{x}}(t)      & = & \tilde{f}(\mathbf{x}) + \tilde{\mathbf{c}}(t) \mathrm{, with} \\
    \tilde{f}(\mathbf{x})    & = & 1/\tau(-\mathbf{x}(t) + \hat{\mathbf{\Omega}}f(\mathbf{x}(t))+ \epsilon) \mathrm{ and} \\
    \tilde{\mathbf{c}}(t)    & = & 1/\tau(\hat{\mathbf{F}}\mathbf{c}(t) + \mathbf{b}) \mathrm{,}
\end{IEEEeqnarray*}
implying that an \gls{ebn} can be trained to implement the dynamics of \textit{any} given \gls{rnn} obeying the
dynamics of Eq.~\ref{eq:rate_dyn}.

Let us now restate the dynamics of the spiking network with adapted notation for ease of understanding:
\begin{equation*}
  \dot{V}(t) = -\lambda V(t) + \mathbf{F}\tilde{\mathbf{c}}(t) - \Omega^{\mathbf{f}}\mathbf{o}(t) + \Omega^{\mathbf{s}}\mathbf{o}(t) + k\mathbf{D}^T\mathbf{e}(t)
\end{equation*}
And let us assume that we have trained an \gls{rnn} receiving inputs $\mathbf{c}$ to successfully produce a good approximation $\hat{\mathbf{y}} = \hat{\mathbf{D}}\mathbf{x}$ of the target~$\mathbf{y}$, so that $\hat{\mathbf{y}} \approx \mathbf{y}$.

We now see that we can train a network of spiking neurons to encode the \textit{dynamics} $\mathbf{x}$ of the \gls{rnn}, by giving the spiking network input $\tilde{\textbf{c}}(t)=1/\tau(\hat{\mathbf{F}}\textbf{c}(t) + \textbf{b})$. This makes the \textit{dynamics} of the \gls{rnn} the new target of the spiking network: $\tilde{\mathbf{x}}=\mathbf{Dr}$,
so that $\tilde{\mathbf{x}} \approx \mathbf{x}$. After the recurrent weights $\mathbf{\Omega^s}$ have been learned to implement a network that encodes the rate-network dynamics, classification
can be performed by the simple computation $y(t) = \hat{\mathbf{D}}\mathbf{D}\textbf{r}(t)$, where $\hat{\mathbf{D}}$ are the rate-network read-out weights, $\mathbf{D}$ are the spiking read-out weights and $\textbf{r}(t)$ are the filtered spike trains at time $t$ of the spiking network.

% \subsection*{Rate network}
% Our method aims to implement a non-linear dynamical system, described by an \gls{rnn} trained using \gls{bptt}. The \gls{rnn} has $\hat{N}$ units that follow the dynamics described by
% \begin{equation*}
%   \tau_j \dot{x}_j = -x_j + \hat{\mathbf{F}}c_j + \hat{\mathbf{\Omega}}f(x)_j + b_j + \epsilon_j
% \end{equation*}
% where $f(\cdot)$ is for example $\textrm{tanh}(\cdot)$ and $\epsilon$ is a noise term, set to 0.01 during training and to 0 during inference. The trainable parameters in this network are the time constants $\tau$; the encoding and recurrent weights $\hat{\mathbf{F}}$ and $\hat{\Omega}$; and the biases $b$.

\subsection*{Simulation and Learning}
To simulate our network, we used a Jax\cite{jax2018github} implementation. In all experiments, we used a simple Euler integration method with a time-step of \SI{1}{\milli\second}.

To ensure that during learning the reconstructed dynamics closely follow the target dynamics, the term $\textbf{e} = \textbf{x} - \hat{\textbf{x}}$ is computed and fed back into the network using the feedforward weights  $I_{k\mathbf{D}^T\textbf{e}} = k \mathbf{Fe}$.
The error was then used to compute the update of the slow recurrent weights according to $\dot{\Omega}^{\textbf{s}} = \eta \textbf{r}(t) (\mathbf{D}^T\textbf{e}(t))^T$.
In the batched-update version, the filtered spike trains $\textbf{r}(t)$ and errors $\textbf{e}(t)$ are then collected in two matrices -- $\mathbf{R} \in \mathcal{R}^{N \times T}$ and $\mathbf{E} \in \mathcal{R}^{\hat{N} \times T}$ -- which are then used to compute the update $\dot{\Omega}^{\textbf{s}} = \mathbf{R} (\mathbf{FE})^T$.
In light of the constraint of some neuromorphic chips that reset potentials can not be calibrated on a per-neuron basis, we did not permit updates to the diagonal of $\Omega^\mathbf{s}$, which was always set to zero.
To furthermore avoid uneven update magnitudes due to the richness of the input (some input signals have only short periods of sound, while others -- typically the negative samples -- have long ones) we normalised $\dot{\Omega}^{\textbf{s}}$ by $({\sum_i \sum_j{\dot{\Omega}^{\textbf{s}}_{i,j}}})/{N^2}$.
Note that this only applies to the batched update case, which we did not use in the experiments.

Finally, the slow recurrent weights are updated according to
\begin{equation*}
    \Omega^\mathbf{s} = \Omega^\mathbf{s} + \eta \dot{\Omega}^{\textbf{s}}
\end{equation*}
where $\eta = \num{1e-5}$ for the audio task and $\eta = \num{5e-6}$ for the temporal XOR task.

\subsection*{Temporal XOR task}
In this experiment, we trained an \gls{rnn} with $\hat{N} = 64$ units to implement a classifier for the temporal XOR task.
This task consists of a one-dimensional temporal signal comprising two sequentially-presented inputs of varying sign to be classified according to the logical operator XOR.
The result of the XOR is signalled by an output (and target) signal which is zero until the withdrawal of the second input, after which the output should go to $\pm 1$ to indicate the result of the XOR.
For example, a signal with two positive -- or two negative -- bumps should be classified as a negative sample and a signal with two bumps that have opposite signs should be classified as a positive example.
The temporal nature of the signal makes this task non-trivial, as the network of spiking neurons needs to keep track of the first half of the signal in order to make the correct decision.
We then trained a spiking network of $N = 320$ neurons to implement the dynamical system described by the \gls{rnn}. The parameters of the spiking network are given in Table~\ref{tab:temporal_xor_parameters}.

\begin{table}[h]
\centering
\caption{\textbf{Parameters for the temporal XOR task.}}
 \begin{tabular}{ll}
  \toprule
  \textit{Parameter}            &   \textit{Value} \\
  $N_c$                         & 1  \\
  $\hat{N}$                     & 64  \\
  $N$                           & 320  \\
  $\tau_{\mathrm{mem}}$     & \SI{50}{\milli\second}  \\
  $\tau_{\mathrm{fast}}$    & --  \\
  $\tau_{\mathrm{slow}}$    & \SI{70}{\milli\second}  \\ \bottomrule
  \end{tabular}
  \label{tab:temporal_xor_parameters}
\end{table}

\subsection*{Network architectures}
We investigated the robustness to simulated noise for four different learning paradigms, including the FORCE method, \gls{bptt} and reservoir computing.
We implemented FORCE, as well as \gls{bptt} using Jax\cite{jax2018github} as part of Rockpool\cite{rockpool}.
To ensure comparability, we chose most of the parameters such as network size and input dimensionality to be the same across different architectures. Table~\ref{tab:mismatch_network_parameters} summarises the parameters used for each architecture.

\begin{table}[h]
\centering
\caption{\textbf{Network architecture parameters used during noise robustness experiments.} *: Subject to modification during training.}
 \begin{tabular}{lllll}
  \toprule
  \textit{Parameter}        & \textit{FORCE} & \textit{\gls{bptt}} & \textit{Reservoir} & \textit{ADS Network} \\
  $N_c$                     & 16 & 16 & 16 & 16 \\
  $\hat{N}$                 & 128 & -- & -- & 128 \\
  $N$                       & 768 & 768 & 768 & 768 \\
  $\tau_{\mathrm{mem}}$ & \SI{10}{\milli\second} & \SI{50}{\milli\second}* & $\mathcal{U}[1e-4,0.112]$ & \SI{50}{\milli\second} \\
  $\alpha$                  & 0.00001 & -- & -- & --\\
  $\tau_{\mathrm{syn}}$ & \SI{20}{\milli\second} & \SI{70}{\milli\second}* & $\mathcal{U}[1e-4,0.112]$  & \SI{70}{\milli\second}/\SI{1}{\milli\second} \\ \bottomrule
  \end{tabular}
  \label{tab:mismatch_network_parameters}
\end{table}

\subsection*{Measurements of parameter mismatch}
Using recordings from fabricated mixed-signal neuromorphic chips we measured levels of parameter mismatch (i.e. fixed substrate noise pattern) present in hardware. In particular, for DYNAP™-SE\cite{dynapse}, a neuromorphic processor which emulates LIF neurons with alpha and exponential synaptic response using analog circuits, we measured neuron and synaptic time constants and synaptic weights for individual neuron units, by recording and analysing the voltage traces produced by these circuits. We observed levels of mismatch in the order of 10--20\% for individual parameters, with parameter spread being proportional to the mean parameter value (see Fig \ref{fig:mismatch_distribution}).

Measurements were obtained by capturing traces of membrane voltage of individual silicon neuron circuits of the chip using a program controlled oscilloscope. Even though only the neuron membrane voltage could be observed, manipulations of the neuron and synapse circuit biases allow the measurement of various circuit parameters indirectly.

To measure the neuron time constants, neurons were injected with a square step of constant DC current that leads to a stable membrane potential below the spiking threshold (i.e. membrane leakage is equal to constant input current). After the step input is removed the membrane potential voltage decays to resting state, and the resulting trace was fitted with an exponential decay function to extract the membrane time constant for that neuron.

Refractory periods of neurons were measured by injecting the neurons with sufficient constant DC current to emit spikes. Refractory period was measured as the time between the after-spike membrane potential drop until the membrane potential rose back to the 10\% level of the overall trace amplitude.

Synaptic parameters were observed through neuron circuits by setting the neuron time constants to the shortest possible value (i.e. maximizing membrane leakage) so that the membrane voltage directly followed the shape of excitatory and inhibitory synaptic input currents.
By stimulating synapses with regular spike trains with low enough rate for the pulses not to interact with each other, the resulting trace amplitude for each pulse amplitude was considered a measure of weight, and the pulse decay time specified the synaptic time constant.
Note that the pulse amplitudes should rather be considered as a way of characterizing the relative variability of weights rather their absolute values, as the measurements can only be performed indirectly via the neuron circuit.

%% Supplementary section: Relationship between FOLLOW and Alemi
\newpage
\section*{Relationship between Alemi\cite{alemi_etal_2018_learning} and FOLLOW\cite{Gilra_2017} learning rules}
The learning rule proposed in ref.~\citenum{alemi_etal_2018_learning} closely matches the FOLLOW\cite{Gilra_2017} learning rule for the recurrent updates, when a few assumptions are made.

The learning rule described in ref.~\citenum{alemi_etal_2018_learning} is given by
\begin{equation*}
    \dot{\mathbf{W}}^{\mathrm{slow}} = \Psi(r) (\mathbf{D^T}e)^T
\end{equation*}
where $\Psi(r)=\phi(\mathbf{M}r+\theta)$. If we assume $\Psi(r) = r$, which did not have large impact on functionality as demonstrated, we can rewrite the learning rule to
\begin{equation*}
    \dot{\mathbf{W}}^{\mathrm{slow}} = r (\mathbf{D^T}e)^T
\end{equation*}
The authors use the notation that $W_{i,j}$ is the connection from neuron $i$ to~$j$\cite{alemi_etal_2018_learning}. Since the FOLLOW paper assumes the exact opposite, we rewrite the update rule to
\begin{equation*}
    \dot{\mathbf{W}}^{\mathrm{slow}} = (\mathbf{D^T}e)r^T
\end{equation*}
We now define the error current as
\begin{equation*}
    I^\epsilon = \mathbf{D^T}e = \sum_{\alpha=1}^{\hat{N}}{D_{\alpha,i}e_{\alpha}}
\end{equation*}
We can now write the update to $\mathbf{W}^{\mathrm{slow}}$ as
\begin{equation} \label{eq:alemi_update_rewritten}
    \dot{W}_{i,j} = \sum_{\alpha=1}^{\hat{N}}{D_{\alpha,i}e_{\alpha}} (S_j * \kappa)
\end{equation}
where we replaced $r_j$ with the spike train $S_j$ convolved with the synaptic response kernel $\kappa$.

The recurrent update to $W_{i,j}$ in the FOLLOW scheme is given by
\begin{equation} \label{eq:follow_update}
    \dot{W}_{i,j} = (I_i^\epsilon * \kappa^\epsilon)(S_j * \kappa)
\end{equation}
where $I_i^\epsilon = k \sum_{\alpha=1}^{\hat{N}}{D^{\mathrm{FOLLOW}}_{i,\alpha}e_{\alpha}}$

One can see that the updates \ref{eq:alemi_update_rewritten} and \ref{eq:follow_update} are equivalent under the assumptions $\Psi(r) = r$ and $\kappa^\epsilon = \delta$, where $\delta$ is the unit impulse kernel.

\end{document}